%% file: asc_arxiv.tex
\documentclass[10pt,twocolumn,letterpaper]{article}

\usepackage{cvpr}
\usepackage{times}
\usepackage{epsfig}
\usepackage{graphicx}
\usepackage{amsmath}
\usepackage{amssymb}
\usepackage{booktabs}
\usepackage{multirow}
\usepackage{animate}
\usepackage{inconsolata}
\usepackage[numbers,sort]{natbib}

\usepackage[breaklinks=true,bookmarks=false]{hyperref}

\cvprfinalcopy % *** Uncomment this line for the final submission

\def\cvprPaperID{****} % *** Enter the CVPR Paper ID here

\begin{document}

%%%%%%%%% TITLE
\title{Active Speakers in Context}

\author{ \normalsize Juan Le\'on Alc\'azar$^{1}$, Fabian Caba Heilbron $^{2}$, Long Mai$^{2}$, Federico Perazzi$^{2}$, \\ \normalsize Joon-Young Lee$^{2}$, Pablo Arbel\'aez$^{1}$, and Bernard Ghanem$^{3}$  \\ 
\normalsize $^{1}$Universidad de los Andes, $^{2}$Adobe Research, $^{3}$King Abdullah University of Science and Technology, \\
\small $^{1}$\{\texttt{jc.leon,pa.arbelaez}\}\texttt{@uniandes.edu.co;} 
\small $^{2}$\{\texttt{caba,malong,perazzi,jolee}\}\texttt{@adobe.com};
\small $^{3}$\{\texttt{bernard.ghanem}\}\texttt{@kaust.edu.sa} \\ }

\maketitle
%\thispagestyle{empty}

%%%%%%%%% ABSTRACT
\begin{abstract}
     \vspace{-0.4cm}

   Current methods for active speak
er detection focus on modeling short-term audiovisual information from a single speaker. Although this strategy can be enough for addressing single-speaker scenarios, it prevents accurate detection when the task is to identify who of many candidate speakers are talking. This paper introduces the Active Speaker Context, a novel representation that models relationships between multiple speakers over long time horizons. Our Active Speaker Context is designed to learn pairwise and temporal relations from an structured ensemble of audiovisual observations. Our experiments show that a structured feature ensemble already benefits the active speaker detection performance. Moreover, we find that the proposed Active Speaker Context improves the state-of-the-art on the AVA-ActiveSpeaker dataset achieving a mAP of 87.1\%. We present ablation studies that verify that this result is a direct consequence of our long-term multi-speaker analysis.

   \vspace{-0.65cm}
 \end{abstract}

\input{./sections/1intro}

\input{./sections/2relatedwork}
\input{./sections/3method}
\input{./sections/4results}

%

%%%%%%%%% BODY TEXT
\vspace{-0.2cm}
\section{Conclusion}
\vspace{-0.25cm}
We have introduced a context-aware model for active speaker detection that leverages cues from co-occurring speakers and long-time horizons. We have shown that our method outperforms the state-of-the-art in active speaker detection, and works remarkably well in challenging scenarios when many candidate speakers or only small faces are on-screen. We have mitigated existing drawbacks, and hope our method paves the way towards more accurate active speaker detection. Future explorations include using speaker identities as a supervision source as well as learning to detect faces and their speech attribute jointly.
\\\noindent\textbf{Acknowledgments.} This publication is based on work supported by the King Abdullah University of Science and Technology (KAUST) Office of Sponsored Research (OSR) under Award No. OSR-CRG2017-3405, and by Uniandes-DFG Grant No. P17.853122

\input{sections/6Supplemental}

\newpage
\pagebreak

{\small
\bibliographystyle{ieee_fullname}
\bibliography{egbib}
}

\end{document}

%% file: sections/1intro.tex
\section{Introduction}

Active speaker detection is a multi-modal task that relies on the careful integration of audiovisual information. It aims at identifying active speakers, among a set of possible candidates, by analyzing subtle facial motion patterns and carefully aligning their characteristic speech wave-forms. Although it has a long story in computer vision~\cite{cutler2000look}, and despite its many applications such as speaker diarization or video re-framing, detecting active speakers in-the-wild remains an open problem. Towards that goal, the recently released AVA Active-Speaker benchmark \cite{roth2019ava} provides an adequate experimental framework to study the problem.

Recent approaches for active speaker detection \cite{chung2019naver, zhangmulti} have focused on developing sophisticated 3D convolutional models to fuse local audiovisual patterns that estimate binary labels over short-term sequences. These methods perform well on scenarios with a single speaker, but they meet their limits when multiple speakers are present. We argue that this limitation stems from the insufficiency of audio cues to fully solve the problem and from the high ambiguity of visual cues when considered in isolation \cite{roth2019ava}. 

\input{figures/intro_fig.tex}

In a multi-speaker scenario, an appropriate disambiguation strategy would exploit rich, long-term, contextual information extracted from each candidate speaker. Figure \ref{fig:ASCIntro} illustrates the challenges in active speaker detection when there is more than one candidate speaker.  Intuitively, we can fuse information from multiple speakers to disambiguate single speaker predictions. For instance, by analyzing a speaker for an extended period, we can smooth out wrong speech activity predictions coming from short filler words. Likewise, observing multiple candidate speakers, jointly, enables us to understand conversational patterns, \eg that a natural two-speaker conversation consists of an interleaved sequence of the speakers' utterances.

In this paper, we introduce the Active Speaker Context, a novel representation that models long-term interactions between multiple speakers for in-the-wild videos.  Our method estimates active speaker scores by integrating audiovisual cues from every speaker present in a conversation (or scene). It leverages two-stream architectures \cite{chung2016out, chung2017you, chung2019perfect} to encode short-term audiovisual observations, sampled from the speakers in the conversation, thus creating a rich context ensemble. Our experiments indicate that this context, by itself, helps improve accuracy in active speaker detection. Furthermore, we propose to  refine the computed context representation by learning pairwise relationships via self-attention  \cite{vaswani2017attention} and  by modeling the temporal structure with a sequence-to-sequence model \cite{hochreiter1997long}. Our model not only improves the state-of-the-art but also exhibits robust performance for challenging scenarios that contain multiple speakers in the scene.

\noindent\textbf{Contributions.} In this work we design and validate a model that learns audiovisual relationships among multiple speakers. To this end, our work brings two contributions.\footnote{To enable reproducibility and promote future research, code has been made available at: \url{https://github.com/fuankarion/active-speakers-context}}\\
\textbf{(1)} We develop a model that learns non-local relationships between multiple speakers over long timespans (Section \ref{sec:ASCO}). \\
\textbf{(2)} We observe that this model improves the state-of-the-art in the AVA-ActiveSpeaker dataset by $1.6\%$, and that this improvement is a direct result of modeling long-term multi-speaker context (Section \ref{sec:experiemntal_val}).

%% file: figures/intro_fig.tex
\begin{figure}[t!]
    \begin{center}
        \includegraphics[width=0.99\columnwidth]{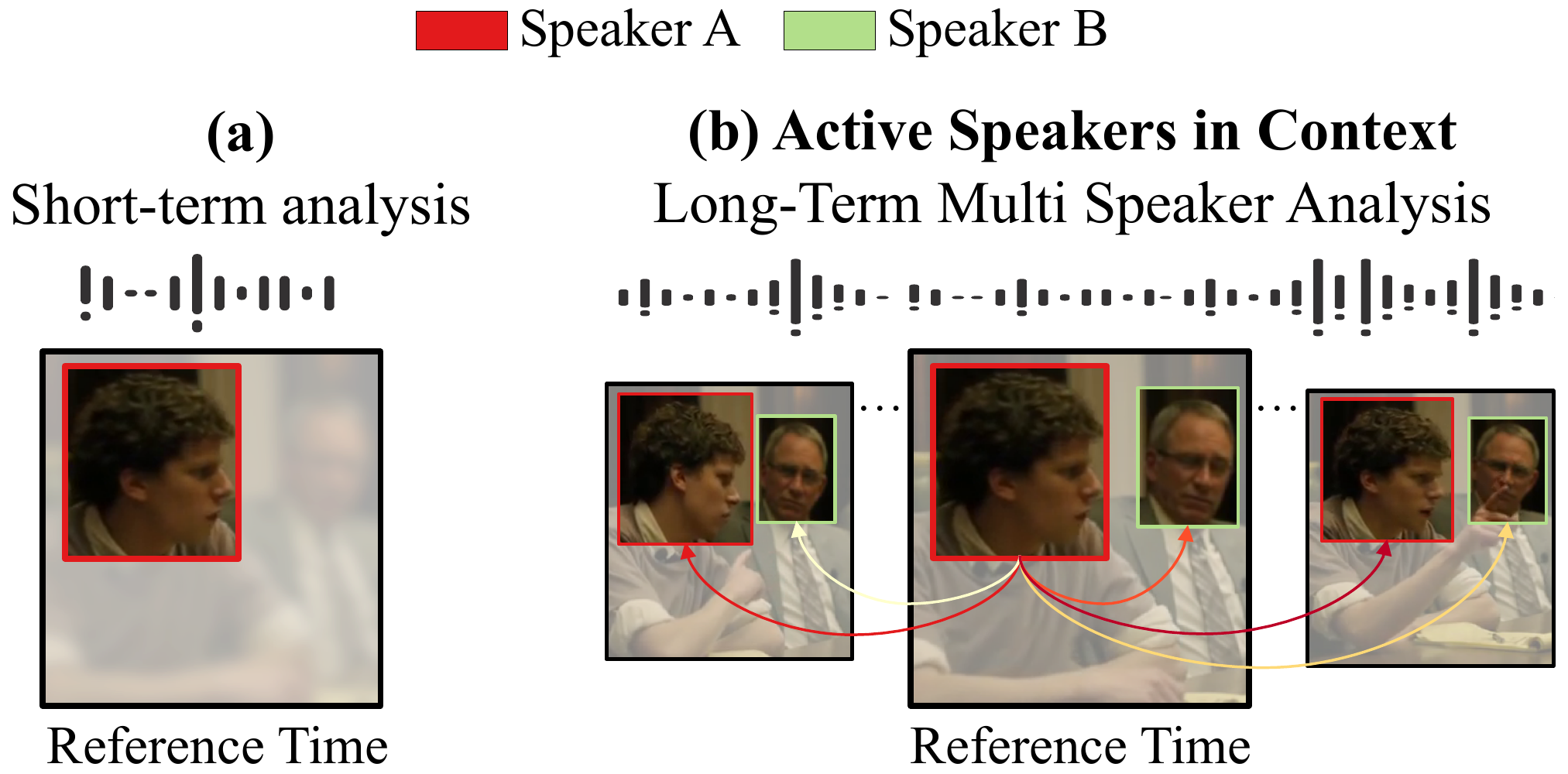}
    \end{center}
    \caption{\textbf{Active Speakers in Context.} Our goal is to identify the active speaker at a reference time. Let us assume we only have access to a short audiovisual sample from a single speaker \textbf{(a)}. By looking at the lips of the speaker, it is hard to tell if he is talking, but the audio indicates that someone at that moment is talking. We have no other option than provide an educated guess. To increase our success prediction chances, let us leverage multi-speaker context \textbf{(b)}. We now observe all speakers in the scene during long-term. From this enriched observation, we can infer two things. First, Speaker B is not talking over the whole sequence, and instead, he is listening to Speaker A. Second, looking at Speaker A (\eg his lips) for the long-term helps us to smooth out local uncertainties. We propose a new representation, the Active Speaker Context, which learns long-term relationships between multiple speakers to make accurate active speaker detections.}
    \label{fig:ASCIntro}
    \vspace{-0.4cm}
\end{figure}

%% file: sections/2relatedwork.tex
\section{Related Work}

Multi-modal learning aims at fusing multiple sources of information to establish a joint representation, which models the problem better than any single source in isolation \cite{ngiam2011multimodal}. In the video domain, a form of modality fusion with growing interest in the scientific community involves the learning of joint audiovisual representations \cite{chakravarty2015s, owens2018audio, nagrani2018seeing, wang2018voicefilter, chung2018voxceleb2, jati2019neural}. This setting includes problems such as person re-identification \cite{nagrani2018learnable, kim2018learning, yadav2018learning}, audio-visual synchronization \cite{chung2016out, chung2017lip}, speaker diarization \cite{zhang2019fully},  bio-metrics \cite{nagrani2018seeing, ravanelli2018speaker}, and audio-visual source separation \cite{chakravarty2015s, owens2018audio, nagrani2018seeing, wang2018voicefilter, chung2018voxceleb2, jati2019neural}.  Active speaker detection, the problem studied in this paper, is an specific instance of audiovisual source separation, in which the sources are persons in a video (candidate speakers), and the goal is to assign a segment of speech to an active speaker, or none of the available sources.

\input{figures/overview_fig.tex}

Several studies have paved the way for enabling active speaker detection using audiovisual cues \cite{cutler2000look, chakravarty2015s, chakravarty2016active, chung2016out}. Cutler and Davis pioneered the research \cite{cutler2000look} in the early 2000s. Their work proposed a time-delayed neural network to learn the audiovisual correlations from speech activity. Alternatively, other methods \cite{saenko2005visualspeech, everingham2009taking} opted for using visual information only, especially the lips motion, to address the task. Recently, rich alignment between audio and visual information has been re-explored with methods that leverage audio as supervision \cite{chakravarty2015s}, or jointly train an audiovisual embedding \cite{chung2016out, nagrani2017voxceleb, chung2018voxceleb2}, that enables more accurate active speaker detection. Although these previous approaches were seminal to the field, the lack of large-scale data for training and benchmark limited their application to in-the-wild active speaker detection in movies or consumer videos.

To overcome the lack of diverse and in-the-wild data, Roth \etal \cite{roth2019ava}, introduced AVA-ActiveSpeaker, a large-scale video dataset devised for the active speaker detection task. With the release of the dataset and its baseline \textendash a two-stream network that learns to detect active speakers within a multi-task setting\textendash ~ a few novel approaches have started to emerge. In the AVA-ActiveSpeaker challenge of 2019, Chung \etal \cite{chung2019naver} improved the core architecture of their previous work \cite{chung2016out} by adding 3D convolutions and leveraging large-scale audiovisual pre-training. The submission of Zhang \etal \cite{zhangmulti} also relied on a hybrid 3D-2D architecture, with large-scale pre-training on two multi-modal datasets \cite{chung2016out, chung2019perfect}. Their method achieved the best performance when the feature embedding was refined using a contrastive loss \cite{hadsell2006dimensionality}. Both approaches improved the representation of a single speaker, but ignored the rich contextual information from co-occurring speaker relationships, and intrinsic temporal structures that emerge from dialogues. 

Our approach starts from the baseline of a two-stream modality fusion but explores an orthogonal research direction. Instead of improving the performance of a short-term architecture, we aim at modeling the conversational context of speakers, \ie to leverage active speaker context from long-term inter-speaker relations. Context modeling has been widely studied in computer vision tasks such as object classification \cite{murphy2004using}, video question answering \cite{zhu2017uncovering}, person re-identification\cite{li2019personid}, or action detection \cite{girdhar2019video, wu2019long}. Despite the existence of many works harnessing context to improve computer vision systems, our model is unique and tailored to detect active speakers accurately. To the best of our knowledge, our work is the first to address the task of active speaker detection in-the-wild using contextual information from multiple speakers. 

%% file: figures/overview_fig.tex
\begin{figure*}[t!]
    \begin{center}
        \includegraphics[width=1\textwidth]{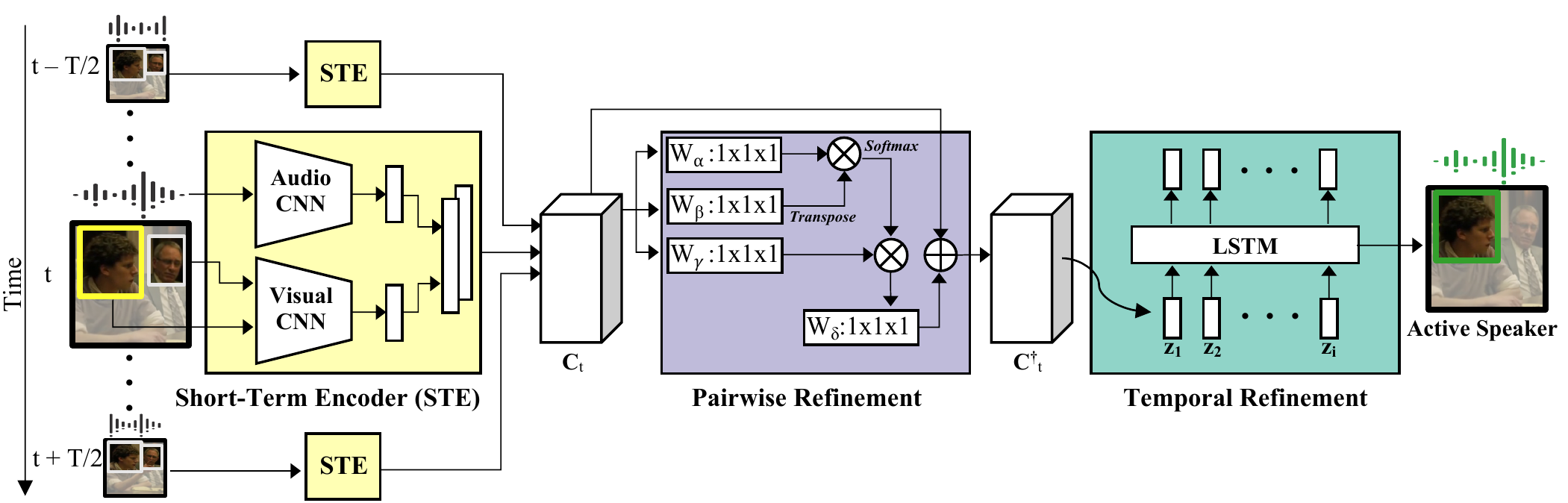}
    \end{center}
    \caption{\textbf{Active Speaker Context.} Our approach first splits the video data into short clips ($\tau$ seconds) composed by a stack of face crops and their associated audio. It encodes each of these clips using a two-stream architecture (Short-Term Encoder) to generate a low-dimensional audiovisual encoding. Then, it stacks the high-level audiovisual features from all the clips and all the speakers sampled within a window of size $T$ ($T > \tau$) centered at a reference time $t$. We denote this stack of features as $\textbf{C}_{t}$. Then, using self-attention, our approach refines the representation by learning a pairwise attention between all elements. Finally, an LSTM mines temporal relationships across the refined features. This final output is our Active Speaker Context, which we use to classify speech activity of a candidate at time $t$.
    }
    \label{fig:ASCOverview}
     \vspace{-0.4cm}
\end{figure*}

%% file: sections/3method.tex
\section{Active Speakers in Context}
\label{sec:ASCO}

This section describes our approach to active speaker detection, which focuses on learning long-term and inter-speaker relationships. At its core, our strategy estimates an active speaker score for an individual face (target face) by analyzing the target itself, the current audio input, and multiple faces detected at the current timestamp. 

Instead of holistically encoding long time horizons and multi-speaker interactions, our model learns relationships following a bottom-up strategy where it first aggregates fine-grained observations (audiovisual clips), and then maps these observations into an embedding that allows the analysis of global relations between clips. Once the individual embeddings have been estimated, we aggregate them into a context-rich representation which we denote as the Active Speaker Ensemble. This ensemble is then refined to explicitly model pairwise relationships, and to explicitly model long-term structures over the clips, we name this refined ensemble the Active Speaker Context. Figure \ref{fig:ASCOverview} presents an overview of our approach.

\subsection{Aggregating Local Video Information}
\vspace{-0.2cm}
\label{sec:method:context-build}
Our proposal begins by analyzing audiovisual information from short video clips. The visual information is a stack of $k$ consecutive face crops \footnote{Our method leverages pre-computed face tracks (consecutive face crops) at training and testing time.} sampled from a time interval $\tau$. The audio information is the raw wave-form sampled over the same $\tau$ interval. We refer to these clips as a tuples $c_{s,\tau}=\{v_{s}, a_{\tau}\}$, where $v_{s}$ is a crop stack of a speaker $s$, and $a_{\tau}$ is the corresponding audio. For every clip $c_{s,\tau}$ in a video sequence, we compute an embedding $\mathbf{u}_{s, \tau}$ using a short-term encoder $\Phi(c_{s,\tau})$ whose role is twofold. First, it creates a low-dimensional representation that fuses the audiovisual information. Second, it ensures that the embedded representation is discriminative enough for the active speaker detection task.

\vspace{-0.4cm}
\paragraph{Short-term Encoder ($\Phi$).} Following recent works \cite{chung2017you, zhangmulti, roth2019ava}, we approximate $\Phi$ by means of a two-stream convolutional architecture. Instead of using compute-intensive 3D convolutions as in \cite{chung2019naver, zhangmulti}, we opt for 2D convolutions in both streams. The visual stream takes as input a tensor $\mathbf{v} \in \mathbb{R}^{H \times W \times (3 k)}$, where $H$ and $W$ are the width and height of $k$ face crops. On the audio stream, we convert the raw audio waveform into a Mel-spectrogram represented as $\mathbf{a} \in \mathbb{R}^{Q \times P}$, where Q and P depend on the length of the interval $\tau$. On a forward pass the visual sub-network estimates a visual embedding $\mathbf{u_v} \in \mathbb{R}^{d_v}$, while the audio sub-network computes an audio embedding $\mathbf{u_a} \in \mathbb{R}^{d_a}$. We compose an audiovisual feature embedding $\mathbf{u} \in \mathbb{R}^{d}$ by concatenating the output embedding of each stream.

\vspace{-0.4cm}
\paragraph{Structured Context Ensemble.} Once the clip features $\mathbf{u} \in \mathbb{R}^{d}$ have been estimated, we proceed to assemble these features into a set that encodes contextual information. We denote this set as the Active Speaker Ensemble.  To construct this ensemble, we first define a long interval $T$ ($T > \tau$) centered at a reference time $t$, and designate one of the speakers present at $t$ as the reference speaker and every other speaker is designated as context speaker. 

We proceed to compute $\mathbf{u}_{s,\tau}$ for every speaker $s={1,\ldots,S}$ present at $t$ over $L$ different $\tau$ intervals throughout temporal window $T$. This sampling scheme yields a tensor $\mathbf{C}_{t}$ with dimensions $L \times S \times d$, where $S$ is the total number of speakers analyzed. Figure \ref{fig:ASC} contains a detailed example on the sampling process.

\input{figures/building_asc.tex}

We assemble $\mathbf{C}_{t}$ for every possible $t$ in a video. Since temporal structures are critical in the active speaker problem, we strictly preserve the temporal order of the sampled features. As $\mathbf{C}_{t}$ is defined for a reference speaker, we can generate as many ensembles $\mathbf{C}_{t}$ as speakers are present at time $t$. In practice, we always locate the feature set of the reference speaker as the first element along the $S$ axis of $\mathbf{C}_{t}$. Context speakers are randomly stacked along the remaining positions on the $S$ axis. This enables us to directly supervise the label of the reference speaker regardless of the number or order of the context speakers. 

\subsection{Context Refinement}
\vspace{-0.2cm}
\label{sec:method:context-refinement}
After constructing the context ensemble $\mathbf{C}_{t}$, we are left with the task of classifying the speaking activity of the designated reference speaker. A naive approach would fine-tune a fully-connected layer over $\mathbf{C}_{t}$ with binary output classes \ie speaking and silent. Although such a model already leverages global information beyond clips, we found that it tends not to encode useful relationships between speakers and their temporal patterns, which emerge from conversational structures. This limitation inspires us to design our novel Active Speaker Context (ASC) model. ASC consists of two core components. First, it implements a multi-modal self-attention mechanism to establish pairwise interactions between the audiovisual observations on $\mathbf{C}_{t}$. Second, it incorporates a long-term temporal encoder, which exploits temporal structures in conversations. 

\vspace{-0.4cm}
\paragraph{Pairwise Refinement.} We start from the multi-modal context ensemble $\mathbf{C}_{t}$, and model pairwise affinities between observations in $\mathbf{C}_{t}$ regardless of their temporal order or the speaker they belong to. We do this refinement by following a strategy similar to Vaswani \etal \cite{vaswani2017attention}. We compute self-attention over long-term sequences and across an arbitrary number of candidate speakers. 

In practice, we adapt the core idea of pair-wise attention from the non-local framework \cite{wang2018non} to work over multi-modal high-level features, thereby estimating a dense attention map over the full set of clips contained in the sampling window $T$. We avoid using this strategy over low or mid-level features as there is no need to relate distributed information on the spatial or temporal domains of a clip \textit{i.e.} in the active speaker detection task,  meaningful information is tightly localized on the visual (lips region) and audio (speech snippets) domains. 

We implement a self-attention module that first estimates a pairwise affinity matrix $\mathbf{B}$ with dimension $LS\times LS$ and then uses its normalized representation as weights for the input $\mathbf{C}_{t}$ :
\vspace{-0.15cm}

\begin{equation}
   \mathbf{B} = \sigma ((W_{\alpha}*\mathbf{C}_{t}) \cdot (W_{\beta}*\mathbf{C}_{t})^{\top})
   \label{eq:attention}
\end{equation}

\vspace{-0.3cm}

\begin{equation}
    \mathbf{C}^{\dagger}_{t} =  W_{\delta} *( \mathbf{B} \cdot (W_{\gamma}*\mathbf{C}_{t}))+\mathbf{C}_{t}
     \label{eq:c_tilde}
\end{equation}

Where $\sigma$ is a softmax operation, $ \{W_{\alpha}, W_{\beta}, W_{\gamma}, W_{\delta}\}$ are learnable $1\times1\times1$ weights that adapt the channel dimensions as needed, and the second term in Equation \ref{eq:c_tilde}  ($+\mathbf{C}_{t}$) denotes a residual connection. The output  $\mathbf{C}^{\dagger}_{t}$ is a tensor with identical dimensions as the input $\mathbf{C}_{t}$ ($L \times S \times d$), but it now encodes the pairwise relationships.

\vspace{-0.4cm}
\paragraph{Temporal Refinement.} 
The goal of this long-term pooling step is two-fold. First, to refine the weighted features in $ \mathbf{C}^{\dagger}_{t}$ by directly attending to their temporal structure. 
%which are more relevant for classifying active speakers. 
Second, to reduce the dimensionality of the final embedding to $d'$ ($d > d'$), allowing us to use a smaller fully-connected prediction layer. Given the inherent sequential structure of the task, we implement this refinement using an LSTM  model \cite{hochreiter1997long}. We cast its input by squeezing the speaker and time dimension of $\mathbf{C}^{\dagger}_{t}$ into $(L \times S) \times d$; thus we input the LSTM time steps  $t_i \in \{1,\ldots,L \times S\}$, with a feature vector $\mathbf{z}_i \in \mathbb{R}^{d}$. In practice, we use a single uni-directional LSTM unit with $d'=128$, and keep the LSTM memory as it passes over the sequence. Thus, we create a sequence-to-sequence mapping between tensor $\mathbf{C}^{\dagger}_{t} \in \mathbb{R}^{ (L \times S) \times d}$ and a our final Active Speaker Context representation  $\mathbf{ASC}_{t} \in \mathbb{R}^{ (L \times S) \times d'}$.

Our final step consists of estimating the presence of an active speaker given $\mathbf{ASC}_{t}$. We resort to a simple fully-connected layer with binary output (active speaker and silent). We establish the final confidence score using a softmax operator over the outputs and select the value of the speaking class.

\subsection{Training and Implementation Details}
\vspace{-0.2cm}
\label{sec:method:training}
We use a two-stream (visual and audio) convolutional encoder based on the Resnet-18 architecture \cite{he2016deep} for the Short-Term Feature extraction (STE). Following \cite{roth2019ava}, we re-purpose the video stream to accept a stack of $N$ face crops by replicating the weights on the input layer $N$ times. The audio stream input is a Mel-spectrogram calculated from an audio snippet, which exactly matches the time interval covered by the visual stack. Since Mel-spectrograms are 2D tensors, we re-purpose the input of the audio stream to accept a $L \times P \times 1$ tensor by averaging channel-specific weights at the input layer. 
 
\vspace{-0.4cm}
\paragraph{Training the Short-term Encoder}
We train the STE using the Pytorch library \cite{paszke2017automatic} for 100 epochs. We choose the ADAM optimizer \cite{kingma2014adam} with an initial learning rate of $3\times 10^{-4}$ and learning rate annealing $\gamma =0.1$ every 40 epochs. We resize every face crop to 124$\times$124 and perform random flipping and corner cropping uniformly along the visual input stack. We drop the large-scale multi-modal pre-training of \cite{chung2019naver}, in favor of standard Imagenet \cite{deng2009imagenet} pre-training for the initialization.

Since we want to favor the estimation of discriminative features on both streams, we follow the strategy presented by Roth \etal \cite{roth2019ava} and add two auxiliary supervision sources, and place them on top of both streams before the feature fusion operation, this creates two auxiliary loss functions $ \mathcal{L}_{a} , \mathcal{L}_{v}$. Our final loss function is $\mathcal{L} = \mathcal{L}_{av} + \mathcal{L}_{a} + \mathcal{L}_{v}$.  We use the standard Cross-entropy loss for all three terms. 

\vspace{-0.4cm}
\paragraph{Training the Active Speaker Context Model}
We also optimize the ASC using the Pytorch library and the ADAM optimizer with an initial learning rate of $3\times 10^{-6}$ and learning rate annealing $\gamma =0.1$ every 10 epochs. We train the full ASC module from scratch and include batch normalization layers to favor faster convergence \cite{ioffe2015batch}. Similar to the STE, we use Cross-entropy loss to train ASC, but in this scenario, the loss consists of a single term $\mathcal{L}_{av}$.

The ASC processes a fixed number of speakers $S$ to construct $\mathbf{C}_t$. Given that not every reference time $t$ contains the same number of speaker detections, there are three scenarios for $J$ overlapping speakers and an ensemble of size $S$. If $J \geq S$, we randomly sample $S-1$ context speakers (one is already assigned as reference). If $J<S$, we select a reference, and randomly sample (with replacement) $S-1$ context speakers from the remaining $J-1$ speakers. In the extreme case where $J=1$, the reference speaker is replicated $S-1$ times. 

%we randomly sample $S-1$ context speakers to fill $\mathbf{C}_{t}$. If less than $S-1$ speakers are available at time $t$, we randomly replicate existing speakers until $\mathbf{C}_t$ is entirely filled. 

%% file: figures/building_asc.tex
\begin{figure}[t!]
    \begin{center}
        \includegraphics[width=0.5\textwidth]{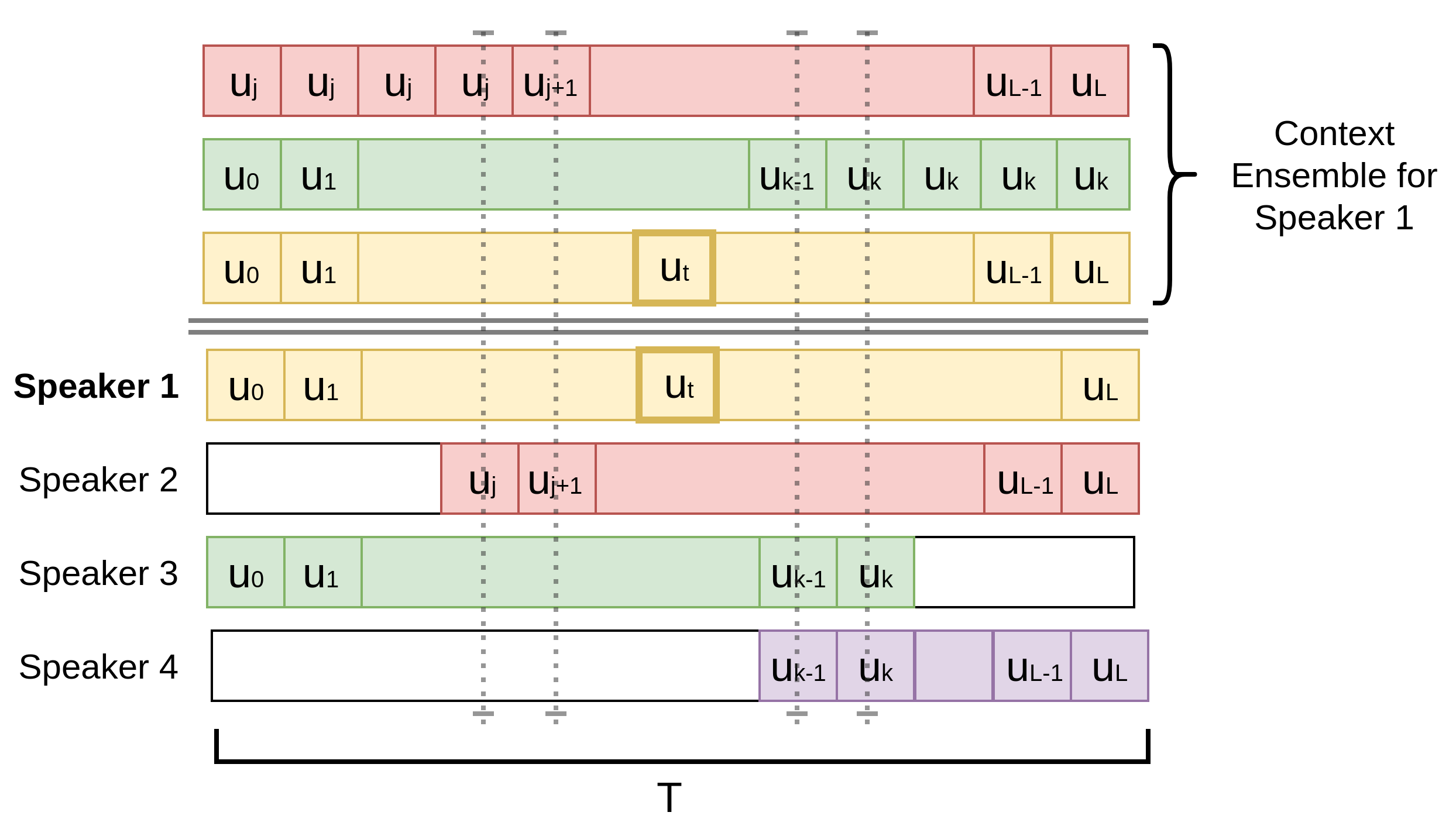}
    \end{center}
    \caption{\textbf{Building Context Tensors.} We build a context ensemble given a reference speaker (Speaker 1 in this example), and a reference time $t$. First, we define a long-term sampling window $T$ containing $L+1$ clips centered at time $t$,  $T=\{0, 1, ..., t, ..., L-1, L\}$. We select as context speakers those that overlap with the reference speaker at $t$ (speakers 2 and 3). Finally, we sample clip-level features $u_{l}$ throughout the whole sampling window $T$ from the reference speaker and all the speakers designated as context. If the temporal span of the speaker does not entirely match the interval T, we pad it with the initial or final speaker features. For instance, Speaker 2 is absent between 0 and $i$, so we pad left with $u_{i}$. Similarly,  for speaker 3, we pad right with $u_{k}$. Notice that, by our definition, Speakers 2 and 3 could switch positions, but Speaker 1 must remain at the bottom of the stack.}
    \label{fig:ASC}
     \vspace{-0.4cm}
\end{figure}

%% file: sections/4results.tex
\section{Experiments}
\label{sec:experiemntal_val}
This section evaluates our method's ability to detect active speakers in untrimmed videos. We conduct the experiments using the large-scale AVA-ActiveSpeaker dataset \cite{roth2019ava}. We divide the experiment analyses into three parts. First, we compare our approach with the existing state-of-the-art approaches. Then, we ablate our method and inspect the contributions of each of its core components. Finally, we do a performance breakdown and analyze success and failure modes.

\vspace{-0.4cm}
\paragraph{AVA-ActiveSpeaker dataset.} The AVA-ActiveSpeaker dataset \cite{roth2019ava} contains $297$ Hollywood movies, with $133$ of those for training, $33$ for validation and $131$ for testing. The dataset provides normalized bounding boxes for 5.3 million faces (2.6M training, 0.76M validation, and 2.0M testing) detected over 15-minute segments from each movie. These detections occur at an approximate rate of 20fps and are manually linked over time to produce face tracks depicting a single identity (actor). Each face detection in the dataset is augmented with a speaking or non-speaking attribute. Thus, the task at inference time is to produce a confidence score that indicates the chance of speaking for each given face detection. In our experiments, we use the dataset official evaluation tool, which computes the mean average precision (mAP) metric over the validation (ground-truth available) and test sets (ground-truth withheld). Unless mentioned otherwise, we evaluate active speaker detection on the AVA-ActiveSpeaker validation subset. 

\vspace{-0.4cm}
\paragraph{Dataset sampling at training time.} As noted by Roth \etal \cite{roth2019ava}, AVA-ActiveSpeaker has a much smaller variability in comparison to natural image datasets with a comparable size. For the training of the STE, we prevent over-fitting by randomly sampling a single clip with $k$ time contiguous crops from every face track instead of densely sampling every possible time contiguos clip of size $k$ in the tracklet. Therefore, our epoch size correlates with the number of face tracks rather than the number of face detections. To train our context refinement models, we use standard dense sampling over the training set, as we do not observe any significant over-fitting in this stage.

\subsection{Comparison with the State-of-the-art}
\vspace{-0.2cm}
\label{sec:experiments:sota}
We compare our method's performance to the state-of-the-art and summarize these results in Table \ref{tab:state_of_the_art}. We set $L=11$ and $S=3$ for the experiment. We report results on the validation and testing subsets. The latter is withheld for the AVA-ActiveSpeaker task in the ActivityNet challenge \cite{caba2015activitynet}. 

\input{tables/state_of_the_art.tex}

We observe that our method outperforms all existing approaches in the validation subset. This result is very favorable as the other methods rely on 3D convolutions and large scale pre-training, while our model relies exclusively on contextual information built from 2D models. The best existing approach, Chung \etal \cite{chung2019naver}, obtains $85.5\%$. Even though their method uses a large-scale multi-modal dataset for pre-training, our context modeling outperforms their solution by $1.6\%$. 

As Table \ref{tab:state_of_the_art} shows, our method achieves competitive results in the testing subset. Even though our model discards 3D convolutions and model ensembles \cite{chung2019naver}, we rank 2nd in the AVA-ActiveSpeaker 2019 Leaderboard\footnote{\url{http://activity-net.org/challenges/2019/evaluation.html}}. The overall results on the AVA-ActiveSpeaker validation and testing subsets validate the effectiveness of our approach. We empirically demonstrate that it improves the state-of-the-art, but a question remains. What makes our approach strong? We answer that question next via ablation studies.

\subsection{Ablation Analysis}
\vspace{-0.2cm}
\label{sec:experiments:ablation}

\paragraph{Does context refinement help?}
We first assess the effectiveness of the core components of our approach. Table \ref{tab:context_impact} compares the performance of the baseline, a two-stream network (No Context) that encodes a single speaker in a short period, a naive context prediction using a single linear layer (Context + No Refinement), and three ablated variants of our method,  two of these variants verify the individual contributions of the two ASC refinement steps (Context + Pairwise Refinement and Context + Temporal Refinement), the third (Context + Pairwise Refinement + MLP) has a two layer perceptron which yields about the same number of parameters as the ASC, it is useful to test if the increased performance derives from the increased size of the network.

\input{tables/context_impact.tex}
While the initial assembly of the context tensor already improves the baseline performance, our results show that context refinement brings complementary gains. That is, the active speaker detection task benefits not only from the presence of additional clip information in the context, but also profits from directly modeling speaker relationships and temporal structures. We observe that our whole context refinement process leads to an average of $4.73\%$ mAP increase over the context tensor and a naive prediction. These results validate our design choice of distilling context via the pairwise and temporal refinement modules.

\vspace{-0.4cm}
\paragraph{Are there alternatives for temporal refinement?}
We now compare our temporal refinement strategy against a baseline strategy for temporal refinement.
During the recent ActivityNet challenge, Chung \etal \cite{chung2019naver} explored the moving average strategy, reporting an increase of $1.3\%$ mAP using a median filter over prediction scores.
A key difference is that \cite{chung2019naver} processes short-term windows ($0.5$s), whereas we consider windows of $2.25$s.
We found that smoothing long temporal windows negatively impacts the performance of our method.
Table \ref{table:temporalwindows} shows that there is a negligible increase ($+0.02\%$) using short temporal averages, and a drastic drop ($-11.64\%$) using long averages. 

\input{tables/temporal_smoothing.tex}

\vspace{-0.4cm}
\paragraph{Does context size matter?}
We continue the ablation by analyzing the influence of context size on the final performance of our method. Table \ref{tab:context_size} summarizes the two dimensions of this analysis, where we vary the  temporal support (\ie vary $L$ from $1$ to $11$ clips), or alter the number of context speakers (\ie vary $S$ from $1$ to $3$ speakers).

\input{tables/context_size.tex}
\input{tables/context_sampling.tex}

Overall, extended temporal contexts and more co-occurring speakers at training time favor the performance of our method. These results indicate that the proposed approach utilizes both types of context to disambiguate predictions for a single speaker. We observe a larger gap in performance when switching between one to two speakers ($1.8\%$ on average) than when switching between 2 and 3 ($0.15\%$ on average). This behavior might be due to the relative scarcity of samples containing more than three speakers at training time. Regarding temporal support, we observe gradual improvements by increasing $L$. However, as soon as $L$ reaches $11$,  we see diminishing returns that seem to be correlated with the average length of face tracks in the training subset. The context size analysis performed here supports our central hypothesis that context from long-time horizons and multiple-speakers is crucial for making accurate active speaker detections.
 
\vspace{-0.4cm}
\paragraph{Does context sampling matter?}
We now evaluate the effect of tempering the temporal structure when constructing $\textbf{C}_{t}$. We also assess the effectiveness of 'in-context' speaker information, \ie we study if sampling 'out-of-context' speakers degrades the performance of our approach.  For the first experiment, we build $\mathbf{C}_{t}$ exactly as outlined in Section \ref{sec:method:training}, but randomly shuffle the temporal sequence of all speakers except clips at reference time $t$. For the second experiment we replace the context speakers with a set of speakers sampled from a random time $t'$ such that $t' \neq t$. We report the results in Table \ref{tab:context_sampling}.

Let us analyze the two sampling distortions one at a time. First, the ablation results highlight the importance of the temporal structure. If such a structure is altered, the effectiveness of our method drops below that of the baseline to $77.8\%$. Second, it is also important to highlight that incorporating out-of-context speakers in our pipeline is worse than using only the reference speaker ($84.5\%$ vs. $87.1\%$). In other words, temporal structure and surrounding speakers provide unique contextual cues that are difficult to replace with random information sampled from a video.

\subsection{Results Analysis}
\vspace{-0.2cm}

\paragraph{Performance Breakdown.} Following recent works \cite{alwassel_2018_detad}, we break down our model's and baseline performances in terms of relevant characteristics of the AVA Active Speaker dataset, namely number of faces and face size, which we present in Figure \ref{fig:analysis:performance-breakdown}. We also analyze the impact of noise in speech and find that both our method and the baseline are fairly robust to altered speech quality;
%we expand this analysis in the \textit{supplementary material}.

\input{figures/performance_breakdown.tex}
\input{figures/attention_fig.tex}

The performance breakdown for the number of faces in Figure \ref{fig:analysis:performance-breakdown} \textbf{(left)} reveals the drawbacks of the baseline approach, and the benefits of ASC. We split the validation frames into three mutually exclusive groups according to the number of faces in the frame. For each group, we compute the mAP of the baseline and our approach. Although both follow a similar trend with performance decreasing as the number of faces increases, our method is more resilient. For instance, in the challenging case of three faces, our method outperforms the baseline by $13.2\%$. 
This gain could be due to our method leverages information from multiple speakers at training time, making it aware of conversational patterns and temporal structures unseen by the baseline.

Dealing with small faces is a challenge for active speaker detection methods \cite{roth2019ava}.
Figure \ref{fig:analysis:performance-breakdown} \textbf{(right)} presents how the baseline and our ASC method are affected by face size.
We divide the validation set into three splits: (S) small faces with width less than $64$ pixels, (M) medium faces with width $64$ and $128$ pixels, and (L) large faces with width more than $128$ pixels. There is a correlation between the performance of active speaker detection and face size. Smaller faces are usually harder to label as active speakers. However, our approach exhibits less performance degradation than the baseline as face size decreases. In the most challenging case, \ie small faces, our method outperforms the baseline by $11.3\%$. We hypothesize that our method aggregates information from larger faces via temporal context, which enhances predictions for small faces.

\vspace{-0.4cm}
\paragraph{Qualitative results.} We analyze the pairwise relations built on the matrix $\mathbf{C}_{t}$ on a model trained with only two speakers. Figure \ref{fig:analysis:qualitative} showcases three sample sequences centered at a reference time $t$, each containing two candidate speakers. We highlight the reference speaker in yellow and represent the attention score with a heat-map growing from light-blue (no attention) to red (the highest-attention).

Overall we notice three interesting patterns. First, sequences labeled as silent generate very sparse activations focusing on a specific timestamp (see top row). We hypothesize that identifying the presence of speech is a much simpler task than detecting the actual active speaker. Therefore, our model reliably decides by only attending a short time span.  Second, for sequences with an active speaker, our pairwise refinement tends to distribute the attention towards a single speaker throughout the temporal window (see the second row). Besides, the attention score tends to have a higher value near the reference time and slowly decades as it approaches the limit of the time interval.  Third, we find many cases in which our model attends to multiple speakers in the scene. This behavior often happens when the facial features of the reference speaker are difficult to observe or highly ambiguous. For example, the reference speaker in the third row is hard to see due to insufficient lighting and face orientation in the scene. Hence, the network attends simultaneously to both the reference and the context speaker.

%% file: tables/state_of_the_art.tex
\begin{table}[t!]
    \small
    \centering
    \begin{tabular}{ l c }
    \toprule
    \textbf{Method}  & \textbf{mAP} \\
    \midrule
    \multicolumn{2}{l}{\textit{Validation subset}} \\
    \textbf{Active Speakers Contex (Ours)}    &  \textbf{87.1}  \\
    Chung \etal (Temporal Convolutions)  \cite{chung2019naver} & 85.5 \\
    Chung \etal (LSTM) \cite{chung2019naver}   &   85.1 \\
    Zhang \etal \cite{zhangmulti} & 84.0 \\
    
    \midrule
    \multicolumn{2}{l}{\textit{ActivityNet Challenge Leaderboard 2019}} \\
    Naver Corporation \cite{chung2019naver} & 87.8 \\
    \textbf{Active Speakers Context (Ours)} &  \textbf{86.7}  \\ 
    University of Chinese Academy of Sciences \cite{zhangmulti} & 83.5 \\
    Google Baseline \cite{roth2019ava} & 82.1 \\
    
    \toprule
    \end{tabular}
    \caption{\textbf{Comparison with the State-of-the-art.} 
    We report the performance of state-of-the-art methods in the AVA Active Speakers validation and testing subsets. Results in the validation set are obtained using the official evaluation tool published by \cite{roth2019ava}, test set metrics are obtained using the the ActivityNet challenge evaluation server. In the validation subset, we improve the performance of previous approaches by $1.6\%$, without using large-scale multimodal pre-training. In the test subset, we achieve $86.7\%$ and rank second in the leaderboard, without using 3D convolutions, sophisticated post-processing heuristics or assembling multiple models.
    }
     \label{tab:state_of_the_art}
     
\end{table}

%% file: tables/context_impact.tex
\begin{table}[t!]
    \small
    \centering
    \renewcommand{\tabcolsep}{1mm}
    \begin{tabular}{l c }
        \toprule
        \textbf{Context \& Refinement} & \textbf{mAP}\\
        \midrule
        No Context & $79.5$ \\
        Context + No Refinement & $84.4$\\
        Context + Pairwise Refinement  & $85.2$  \\
        Context + Pairwise Refinement + MLP  & $85.3$  \\
        Context + Temporal Refinement & $85.7$ \\
        \textbf{ASC} & $\mathbf{87.1}$\\
        \toprule
        
    \end{tabular}
    \caption{\textbf{Effect of context refinement.} We ablate the contributions of our method's core components. We begin with a baseline that does not include any context, which achieves $79.5\%$. Then, by simply leveraging context with a linear prediction layer, we observe a significant boost of $4.9\%$. Additionally, we find that adding pairwise and temporal refinement further improves the performance by $0.8\%$ and $1.3\%$ respectively. The ASC best performance is achieved only if both refinement steps are included.}
    \label{tab:context_impact}
    \vspace{-0.25cm}
\end{table}

%% file: tables/temporal_smoothing.tex
\begin{table}[h!]
\footnotesize
\centering
\begin{tabular}{c c c c}
	\toprule
    \textbf{w/o temporal} & \textbf{+ moving} & \textbf{+ moving} & \textbf{+ temporal} \\
    \textbf{refinement} & \textbf{average (0.5s)} & \textbf{average (2.25s)} & \textbf{refinement} \\
 	 \midrule
	85.21\% & +0.02\%  & -11.64\% &  +1.9\% \\ 
	\toprule
\end{tabular}
\caption{\small \textbf{Moving average vs. temporal refinement (mAP).} We observe only marginal benefits when replacing the proposed temporal smoothing step with a moving average, in fact this operation has a large penalty when smoothing longer sampling windows. }
\label{table:temporalwindows}
\vspace{-0.25cm}
\end{table}

%% file: tables/context_size.tex
\begin{table}[h!]
    \small
    \centering
    \begin{tabular}{c c c c}
    \toprule
    \multicolumn{1}{c}{\textbf{Temporal}} & \multicolumn{3}{c}{ \textbf{Number of Speakers ($S$)}} \\
    \multicolumn{1}{c}{\textbf{Support ($L$)} $\downarrow$} & $S=1$ & $S=2$ & $S=3$\\ 
    \midrule
    $L=1$          & $79.5$ & $83.1$ & $82.9$         \\
    $L=3$          & $83.1$ & $84.6$ & $85.0$         \\
    $L=5$          & $84.3$ & $85.8$ & $85.9$       \\ 
    $L=7$          & $84.9$ & $86.4$ & $86.6$       \\ 
    $L=9$          & $85.5$ & $86.7$ & $86.9$           \\ 
    $L=11$         & $85.6$ & $87.0$ & $\mathbf{87.1}$ \\ 

    \toprule
    \end{tabular}
    \caption{\textbf{Impact of context size.} We investigate the effect of different sizes of temporal support and the number of speakers used to construct our context representation. To that end, We report the mAP obtained by different context size configurations. We observe that both types of context play a crucial role at boosting performance. Using our longest temporal support, $L=11$ (2.25 seconds), our method improves the baseline ($L=1$ / $S=1$) by $6.1\%$. Moreover, when combined with context from multiple speakers, \ie $L=11$ / $S=3$, we achieve an additional boost of $1.5\%$ resulting in our best performance of $87.1\%$. In short, our findings reveal the importance of sampling context from long time horizons and multiple speakers.}
    \label{tab:context_size}
    \vspace{-0.25cm}
\end{table}

%% file: tables/context_sampling.tex
\begin{table}[h!]
    \small
    \centering
    \begin{tabular}{c c c c}
        \toprule
        & \multicolumn{3}{c}{\textbf{Sampling Distortion Type}} \\
        & Temporal Order &  Surrounding & None  \\
        \midrule
        \textbf{mAP} & $77.8$ & $84.5$ & $\mathbf{87.1}$ \\
        \toprule    
    
    \end{tabular}
    \caption{\textbf{Effect of context sampling distortion.} We observe that our method looses $2.6\%$ mAP when the context speakers are randomly sampled across the video. It also drastically drops ($-9.3\%$) when the context temporal order is scrambled. These results validate the importance of sampling context for the target face within the right surrounding and preserving its temporal order. 
    }
    \label{tab:context_sampling}
    \vspace{-0.25cm}
\end{table}

%% file: figures/performance_breakdown.tex
\begin{figure}[t!]
    \begin{center}
        \includegraphics[width=0.95\columnwidth]{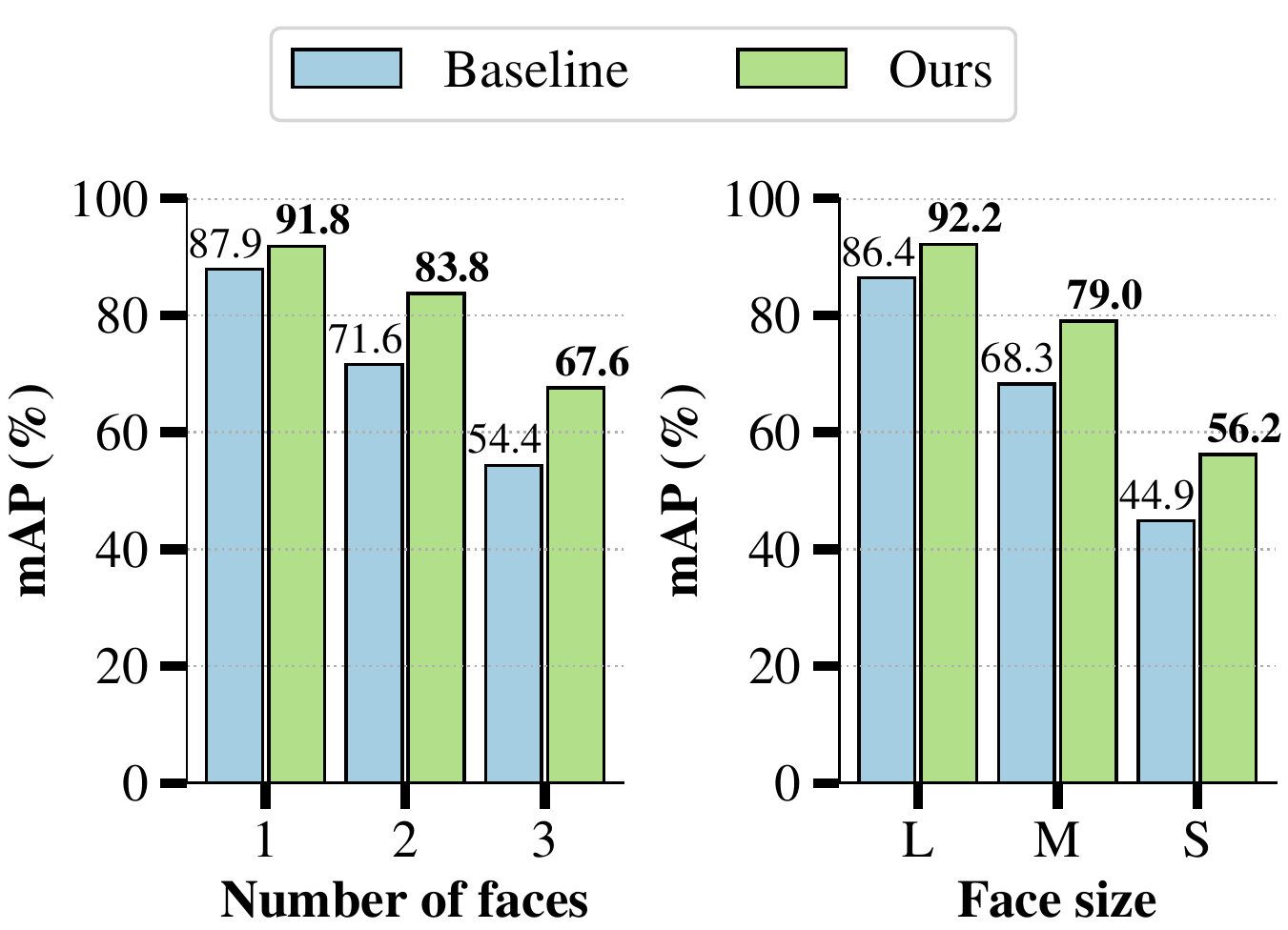}
    \end{center}
    \caption{\textbf{Performance breakdown.} We analyze the performance of the baseline approach (w/o context) and our proposed method (Active Speaker Context) under two different visual characteristics of the samples at inference time: number of faces (\textbf{left}) and face size (\textbf{right}). For the number of faces, we split the dataset into three \textit{exclusive} buckets: one, two, and three faces, which altogether cover $>90\%$ of the dataset. Similarly, we split the dataset into three face sizes: Small (S), Medium (M), Large (L), corresponding to face crops of width $<=64$, $>64$ but $<=128$, and $>128$ pixels, respectively. In all scenarios, we observe that our approach outperforms the baseline, with those gains being more pronounced in challenging scenarios. For instance, when we compare their performance for three ($3$) faces, our method offers a significant boost of $13.2\%$. Moreover, for the hard case of small faces (S), we achieve an improvement of $11.3\%$ over the baseline.}
    \label{fig:analysis:performance-breakdown}
     \vspace{-0.4cm}
\end{figure}

%% file: figures/attention_fig.tex
\begin{figure*}[t!]
    \begin{center}
        \includegraphics[width=0.99\textwidth]{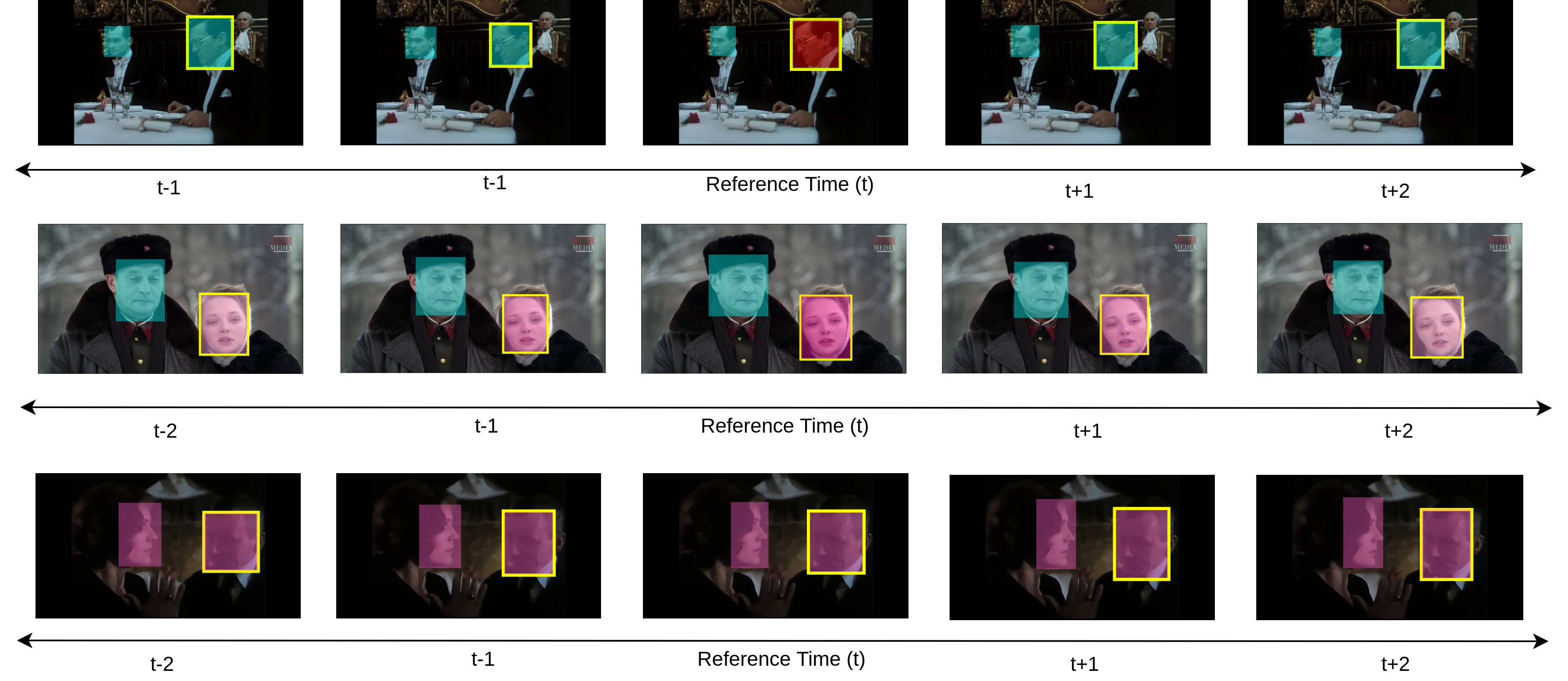}
    \end{center}
    \caption{\textbf{Qualitative results.} The attention within the pairwise refinement step has some characteristic activation patterns. We highlight the reference speaker in a yellow bounding box and represent the attention score with a heat-map growing from light-blue (no attention) to red (the highest-attention). The first row shows a typical activation pattern for two silent speakers. The attention model focuses exclusively on the reference speaker (highlighted in yellow) at the reference time. In the cases where there is an active speaker (second row), the attention concentrates on the reference speaker over an extended time interval. In the third row, the reference speaker is also active, but in this case, his facial gestures are ambiguous; thus, the attention also looks at the context speaker.  
    }
    \label{fig:analysis:qualitative}
    \vspace{-0.4cm}
\end{figure*}

%% file: sections/6Supplemental.tex
\section{Additional Results Analysis}
\label{sec:supplemental}
To complement the experimental validation of section 4, we assess the robustness of our method in presence of audio noise and analyze in detail the pairwise refinement.

\subsection{Performance breakdown}
\begin{figure}[h!]
    \begin{center}
        \includegraphics[width=0.6\columnwidth]{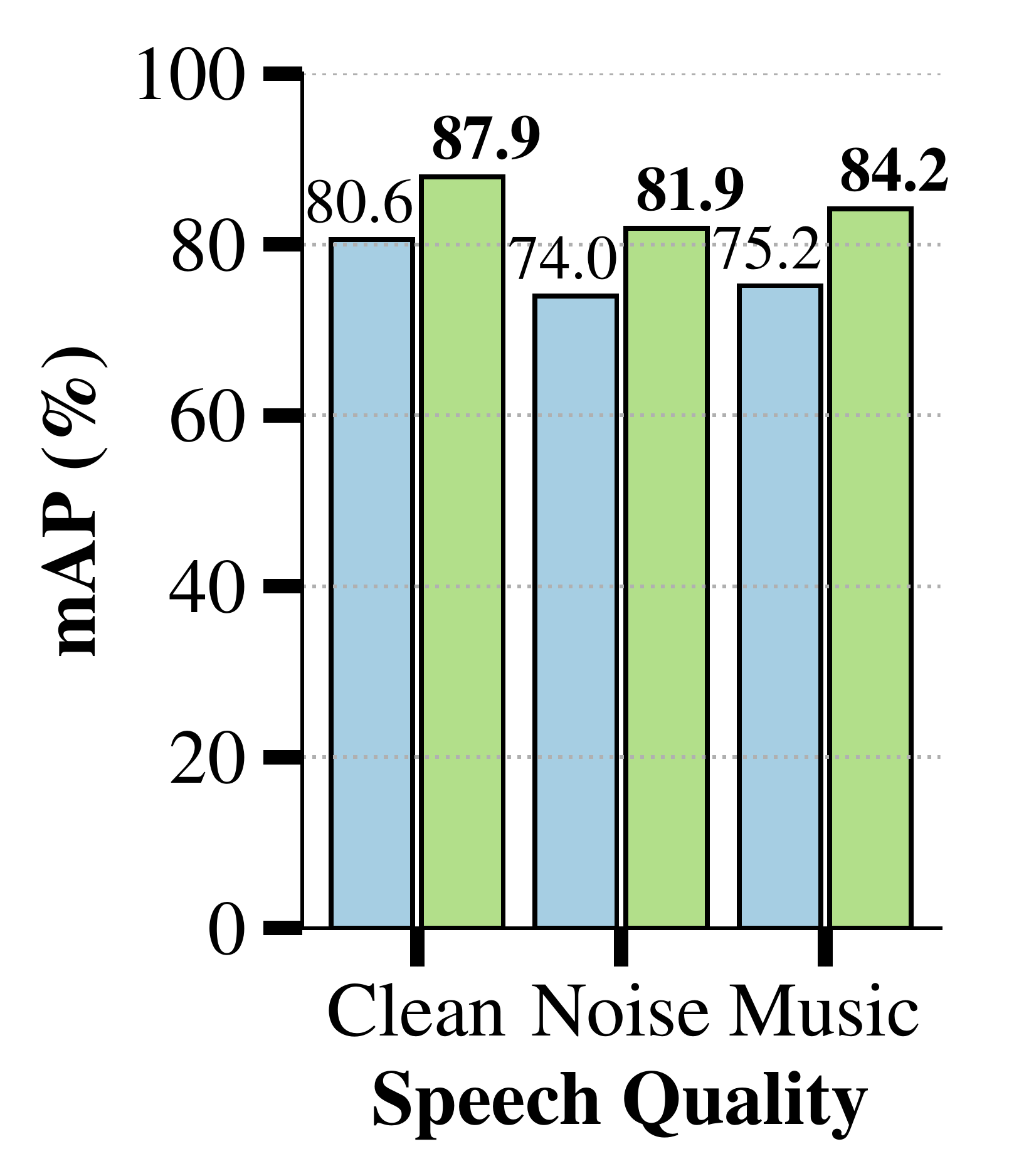}
    \end{center}
    \caption{\textbf{Performance breakdown by speech quality} Out of the two sources of speech quality characterized in AVA-ActiveSpeaker, Music is the less relevant one reducing about 3.7\% mAP for ASC. Noise in speech is a significant source of error, where the proposed approach loses about 6\%. }
    \label{fig:ASCTask}
\end{figure}

\vspace{-0.25cm}
Following the experimental setup of Roth \etal \cite{roth2019ava}, we analyze the performance of our method in the presence of different qualities of speech: Music (common due to Hollywood soundtracks) and other types of noise, figure  \ref{fig:ASCTask} summarizes this results. As outlined by \cite{roth2019ava}, the presence of music in the audio stream is not a significant source of error.  Our ASC model reduces its performance by 3.7\%, and the baseline drops by 5.4\%. We hypothesize the ASC is slightly more resilient to noise as it can look over more extended periods, where clean speech might be present.
 
Noisy speech makes the active speaker detection problem more difficult, where the ASC performance drops by 6\% and the baseline by 6.6\%. Again we believe ASC profits from longer sequences where clean speech patterns are present. Although these error sources have an impact on performance, they are a smaller source of error when compared to the face size and the number of candidate speakers.

\vspace{-0.35cm}

\subsection{Pairwise Refinement Visualization}
\begin{figure*}[h!]
    \begin{center}
        \includegraphics[width=0.99\columnwidth]{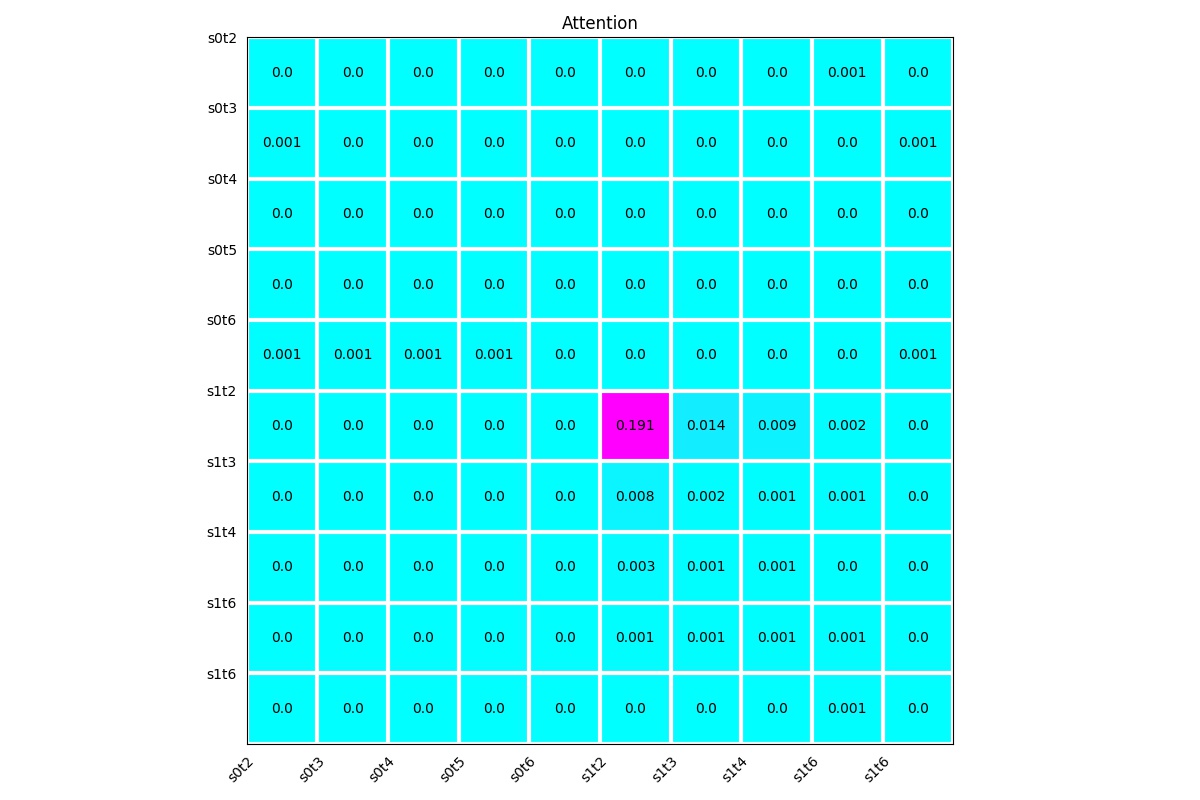}
        \includegraphics[width=0.99\columnwidth]{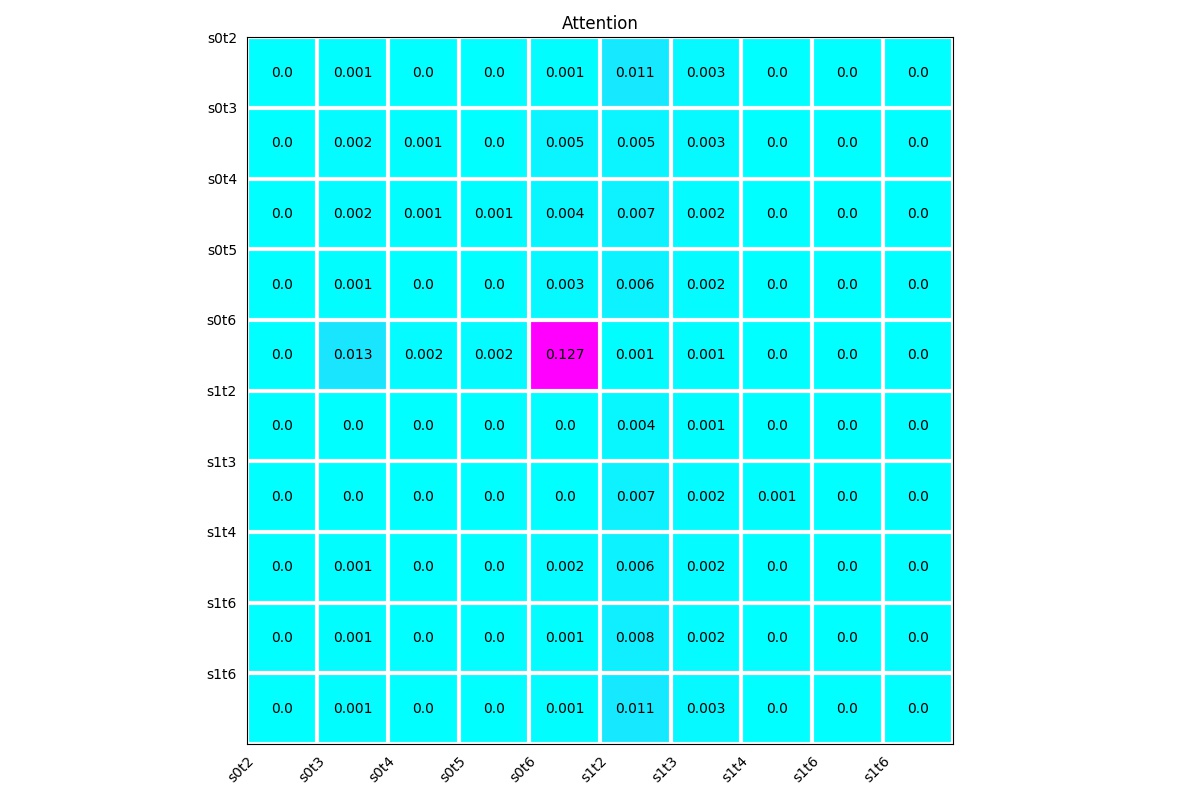} 
    \end{center}
    \caption{\textbf{Activation patterns for silence}  When there are no active speakers, it is common to see activation patterns similar to those pictured above, where very sparse activations are observed over matrix $\mathbf{B}$, we think local cues that indicate silence are strong predictors, thus it is not necessary to look at many.}
    \label{fig:silence}
\end{figure*}

\begin{figure*}[h!]
    \begin{center}
        \includegraphics[width=0.99\columnwidth]{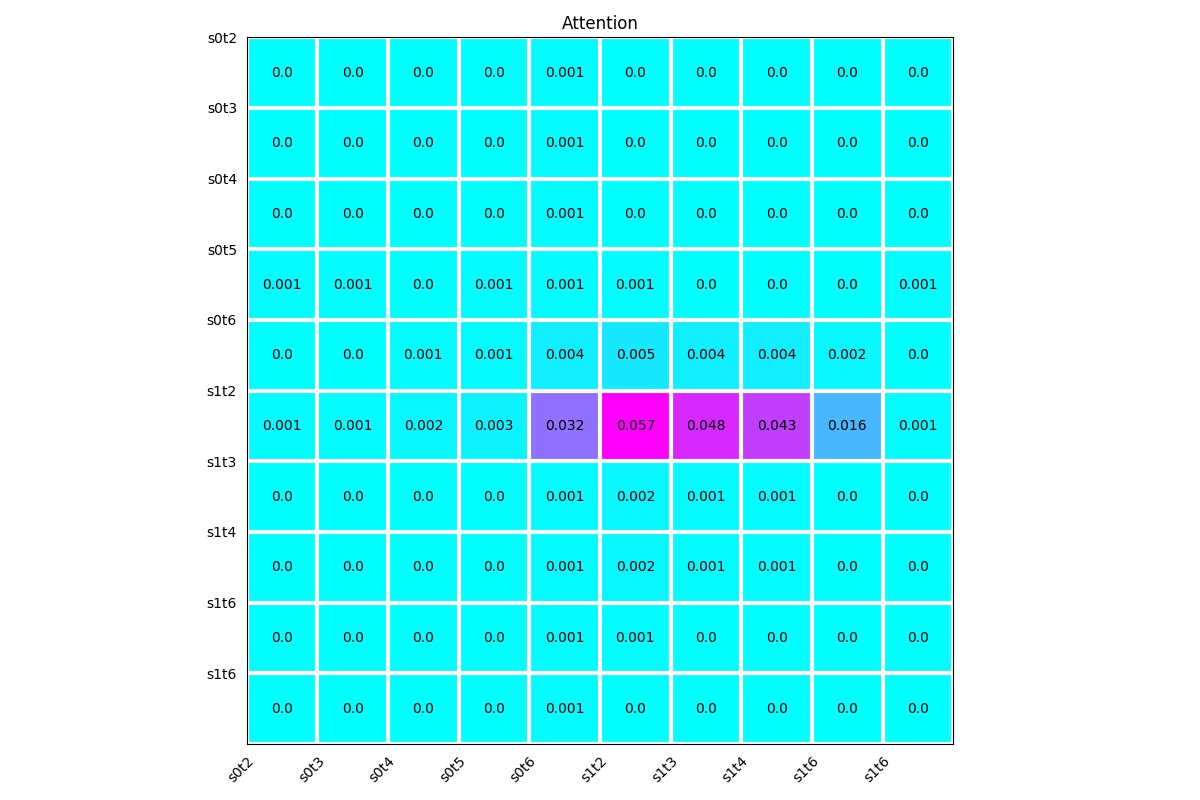}
        \includegraphics[width=0.99\columnwidth]{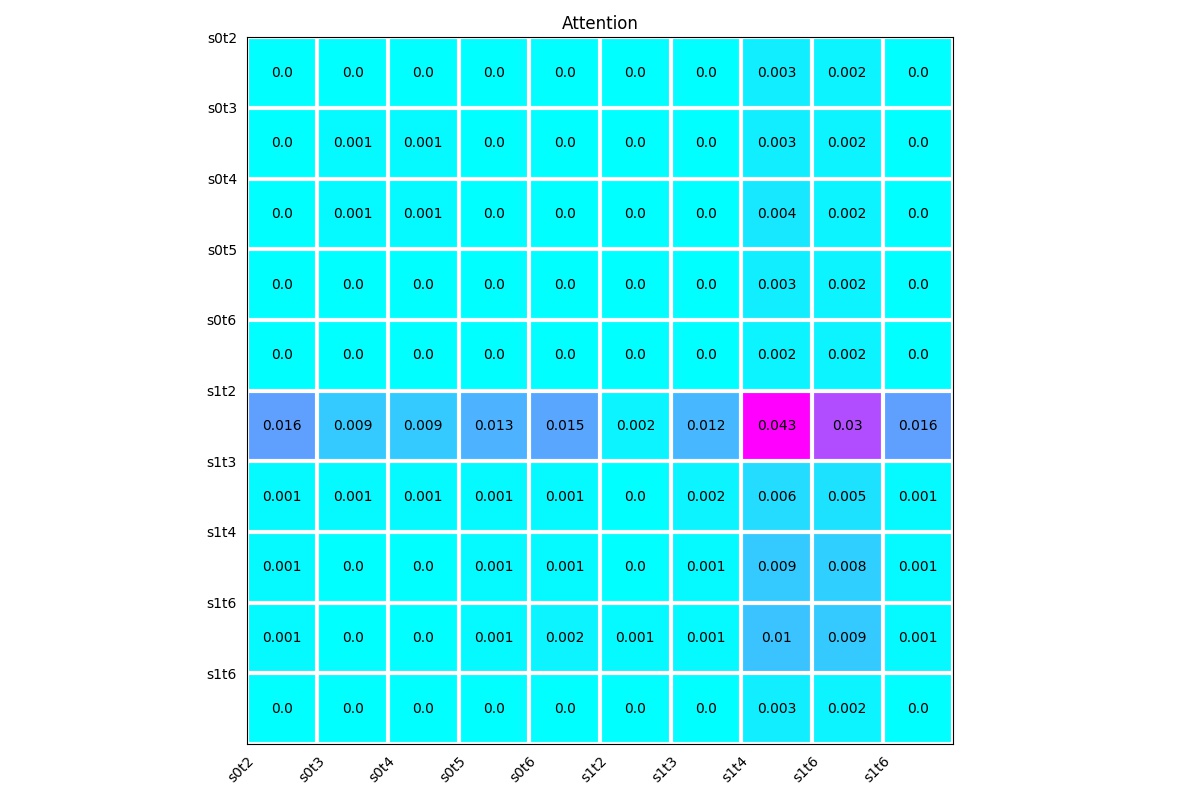} 
    \end{center}
    \caption{\textbf{Activation patterns for an active speaker with no context speakers,}  When there is a single active speaker in the scene we commonly observe activations patterns along a straight line. This patterns suggest the network looks for temporal consistency along several clips in order to predict a high active speaker score.}
    \label{fig:samespeech}
\end{figure*}

\begin{figure*}[h!]
    \begin{center}
        \includegraphics[width=0.99\columnwidth]{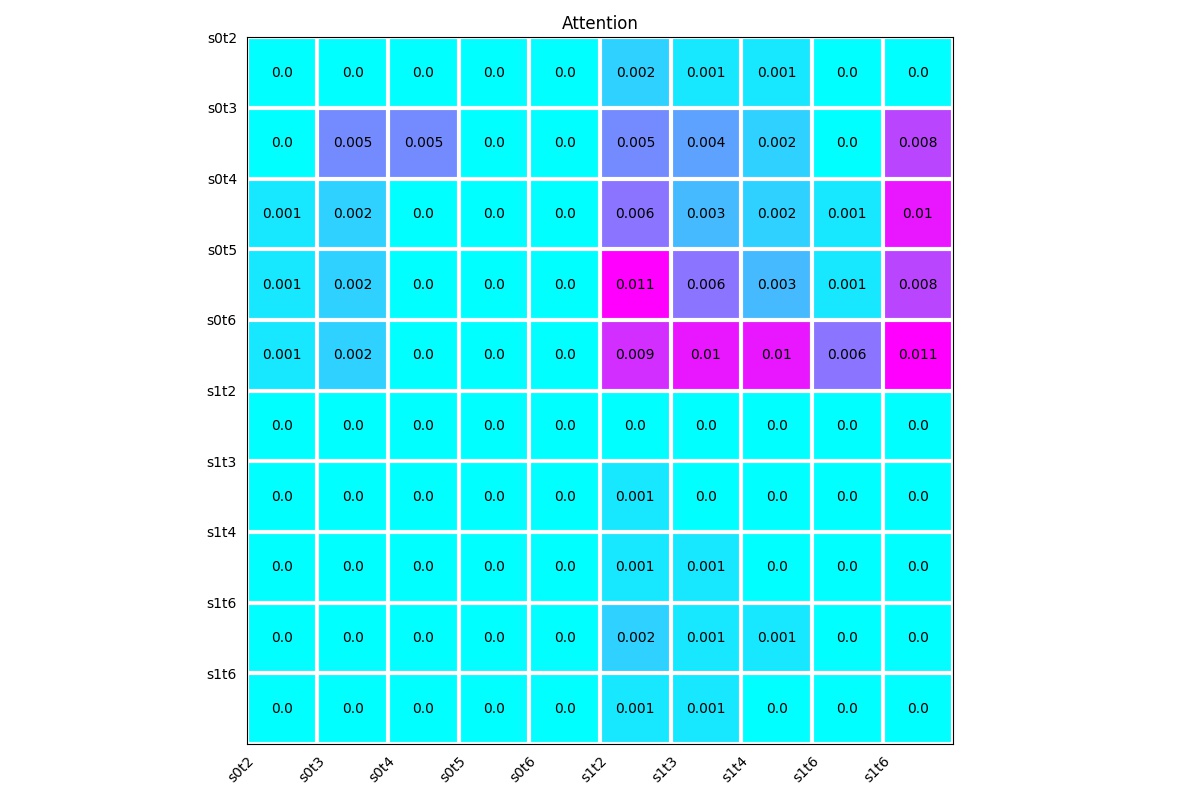}
        \includegraphics[width=0.99\columnwidth]{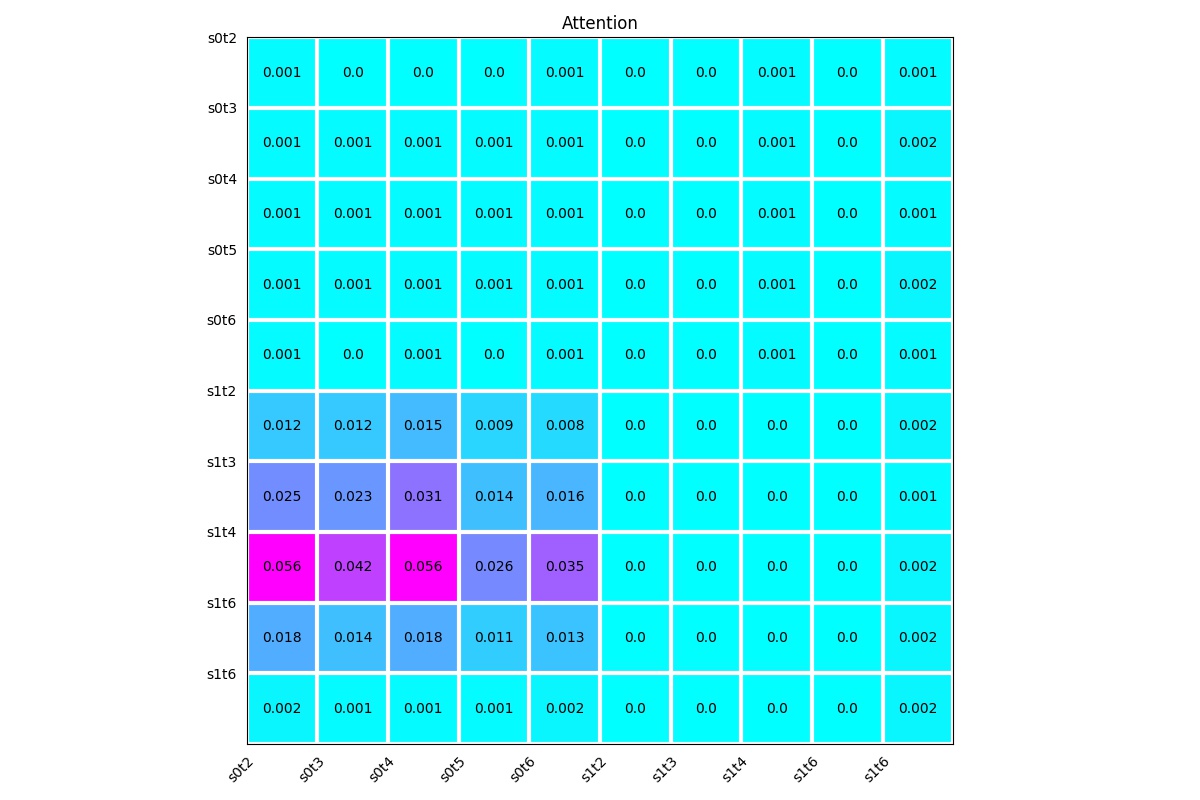} 
    \end{center}
    \caption{\textbf{Activation Patterns for an active speaker with context speaker.} Activation patterns are much more diverse with two speakers, there are some scenarios where the networks focuses on the areas of $\mathbf{B}$ where inter-speaker relations are modelled.}
    \label{fig:diffspeech1}
\end{figure*}

\begin{figure*}[h!]
    \begin{center}
        \includegraphics[width=0.99\columnwidth]{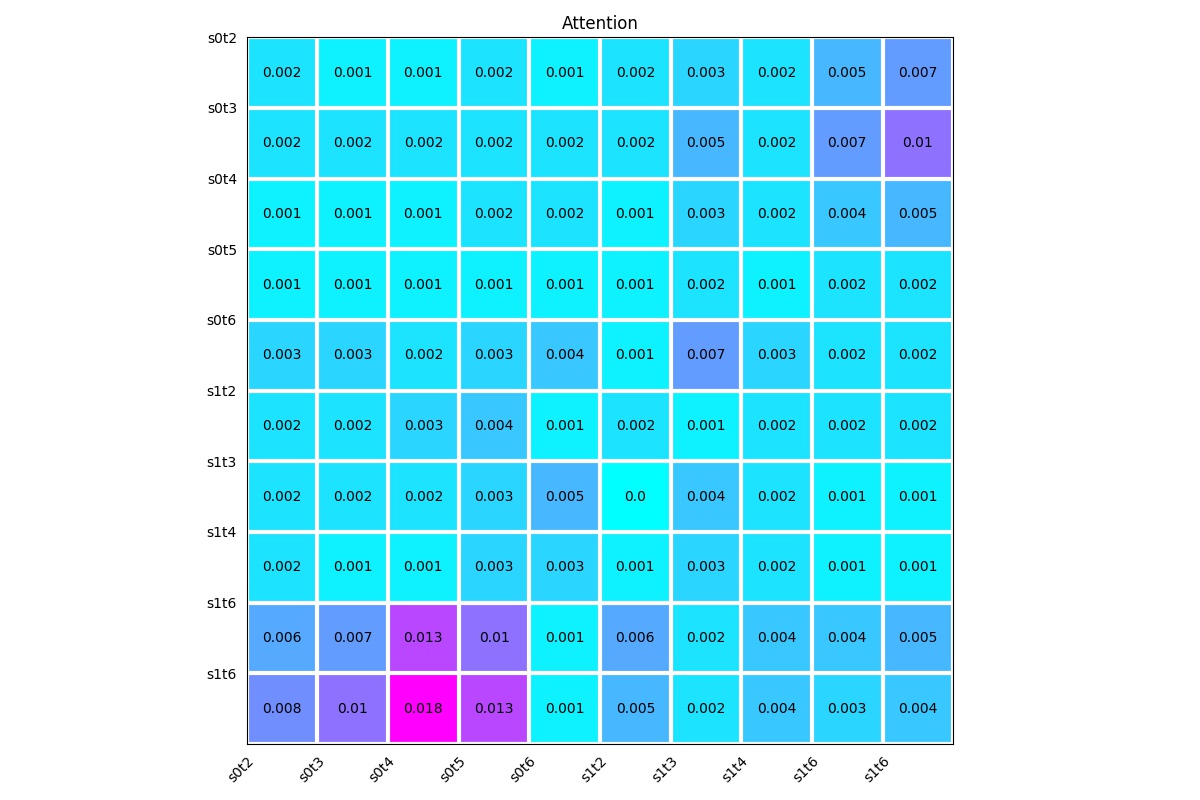}
        \includegraphics[width=0.99\columnwidth]{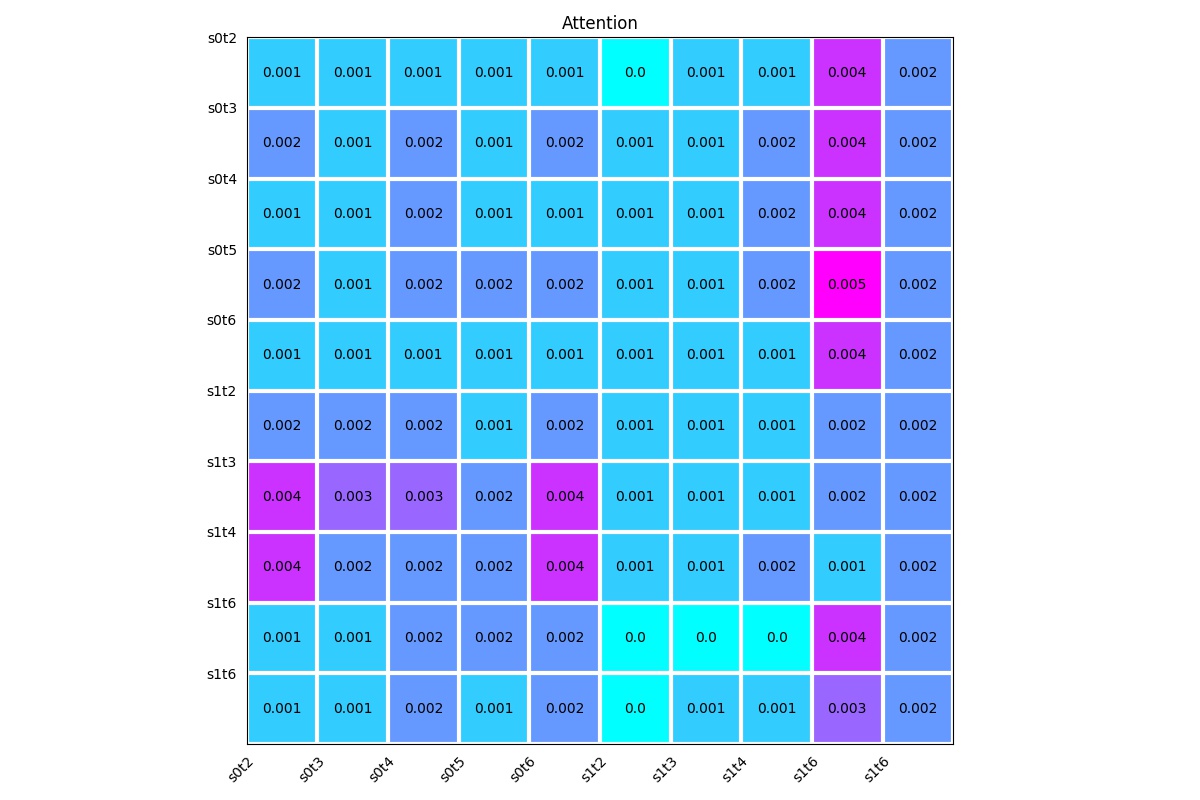} 
    \end{center}
    \caption{\textbf{Activation Patterns for an active speaker with context speaker.}  With two speaker activations patterns are far more diverse than in any other scenario. While the sections of the matrix that attend at inter-speaker relations are highlighted there is also presence of self-attention (see right), and line pattern activations. }
    \label{fig:diffspeech2}
\end{figure*}

\vspace{-0.25cm}
We conclude by visualizing the activation patterns in the pairwise refinement module to better understand the relationships being built amongst the speakers. We use a smaller version of ASC ($S=2$ and $L=9$) than the one in the main paper with only two candidate speakers for easier visualization over nine short-term clips. We plot the activation values of the matrix $\mathbf{B}$ (after the soft-max operation), for the pairwise attention over 9 time steps and 2 speakers. In figures \ref{fig:silence}, \ref{fig:samespeech}, \ref{fig:diffspeech1} and \ref{fig:diffspeech2} we zoom in-between time steps 2 and 6 for simpler plots, and use light-blue for no activation (0.0)  and red for maximum activations (1.0).

We expand the explanation given in Section 4.3 and highlight the activation patterns for 3 common scenarios.

\vspace{-0.35cm}
\paragraph{Silence:} As outlined in the main paper, when both speakers are silent the pair-wise module usually generates sparse activations. This behavior indicates that a single (or few) elements in the Active Speaker Context embedding are enough to reliably estimate the active speaker score. Figure \ref{fig:silence} left and right, show typical activation patterns for this situation, where few elements inside the ensemble have a large attention score. 

Our hypothesis is that local features that correlate to silent scenarios are strong predictors; therefore, inference over the ensemble can be very reliable even if looking at a small number of cues. Overall these are the only scenarios where we find activation values of 1.0. In other cases, we observe that the activation patterns are distributed across the matrix $\mathbf{B}$.

\vspace{-0.35cm}
\paragraph{Active Speaker without context speaker:}  When there is a single candidate speaker in the scene, the ensemble is composed from replicas of the same speaker. In this scenario, we observe some particular activation patterns over $\mathbf{B}$ along straight lines. Such activation patterns indicate that the attention over the ensemble is distributed along the temporal dimension.

We interpret this behavior as the way our pairwise refinement attends to useful temporal information.  Since context and reference speakers have the same information, any line pattern aggregates temporal information of the reference speaker. In other words, the attention mechanism looks for evidence that indicates the speaker is currently talking beyond the reference timestamp. Figure \ref{fig:samespeech} visualizes 2 sample activations for this scenario.

\vspace{-0.35cm}
\paragraph{Active Speaker with context speaker:} The activation patterns for multiple candidate speakers are more complex and harder to characterize as the pairwise attention does not focus on a particular speaker or time-stamp. However, it seems to focus on several time steps establishing relations between the active speaker and the context speaker; notice the activation's on the top right and bottom left of the figure, these areas correspond to inter-speaker attention and contain large activation patterns, while top left and bottom right are self-attention and have less significant activations. Figure \ref{fig:diffspeech1} contains two examples of this activation pattern. 

Overall, the activation pattern for this scenario can be a lot more diverse, and do not always focus on the relation between speakers. In Figure \ref{fig:diffspeech2} we plot some of this scenarios. We will further note that occasionally the activation patterns in this scenario slightly resemble the line activation pattern of a single active speaker.

